\pdfoutput=1

\documentclass[11pt]{article}

\usepackage[final]{acl}

\usepackage{times}
\usepackage{latexsym}

\usepackage[T1]{fontenc}

\usepackage[utf8]{inputenc}

\usepackage{microtype}

\usepackage{inconsolata}

\usepackage{graphicx}

\usepackage{amsmath}
\usepackage{pifont}
\usepackage{enumitem}
\usepackage{booktabs}
\usepackage{multirow}
\usepackage{algorithm}
\usepackage{algorithmic}
\usepackage{hyperref}

\title{Efficient Annotator Reliability Assessment
and 
Sample Weighting
for Knowledge-Based
Misinformation Detection on Social Media
}

\author{Owen Cook, Charlie Grimshaw, Ben Wu, Sophie Dillon, Jack Hicks, Luke Jones, \\ \textbf{Thomas Smith, Matyas Szert} \and \textbf{Xingyi Song} \\
  School of Computer Science,
  The University of Sheffield, \\
  Sheffield, UK \\
  \texttt{\{oscook1, x.song\}@sheffield.ac.uk}
}

\begin{document}
\maketitle

\begin{abstract}
Misinformation spreads rapidly on social media, confusing the truth and targeting potentially vulnerable people. To effectively mitigate the negative impact of misinformation, it must first be accurately detected before applying a mitigation strategy, such as X's community notes, which is currently a manual process. This study takes a knowledge-based approach to misinformation detection, modelling the problem similarly to one of natural language inference. The EffiARA annotation framework is introduced, aiming to utilise inter- and intra-annotator agreement to understand the reliability of each annotator and influence the training of large language models for classification based on annotator reliability. In assessing the EffiARA annotation framework, the Russo-Ukrainian Conflict Knowledge-Based Misinformation Classification Dataset (RUC-MCD) was developed and made publicly available. This study finds that sample weighting using annotator reliability performs the best, utilising both inter- and intra-annotator agreement and soft-label training. The highest classification performance achieved using Llama-3.2-1B was a macro-F1 of 0.757 and 0.740 using TwHIN-BERT-large.
\end{abstract}

\section{Introduction}

Knowledge-based misinformation detection (or fact-checking) is the task of identifying misinformation based on a given piece of knowledge. The knowledge may be either verified misinformation \cite{jiang2021categorising} or true information \cite{thorne2018fever,wadden2020fact}. 

 Datasets \cite{jiang2021categorising, martin2022facter,thorne2018fever} have been introduced to facilitate knowledge-based misinformation detection model training and they are often annotated similarly to Natural Language Inference (NLI) tasks. When the knowledge is verified misinformation, claims can be categorised as \textit{misinformation} if they are \textit{entailed} by the knowledge, a \textit{debunk} if they contradict the knowledge, or as \textit{other} if the claim is \textit{neutral} to the provided knowledge. These datasets often exhibit high levels of annotator disagreement. This high disagreement is mainly due to the high complexity of the task, which introduces subjectivity during annotation.

\begin{figure}[h]
\centering
\includegraphics[width=\columnwidth]{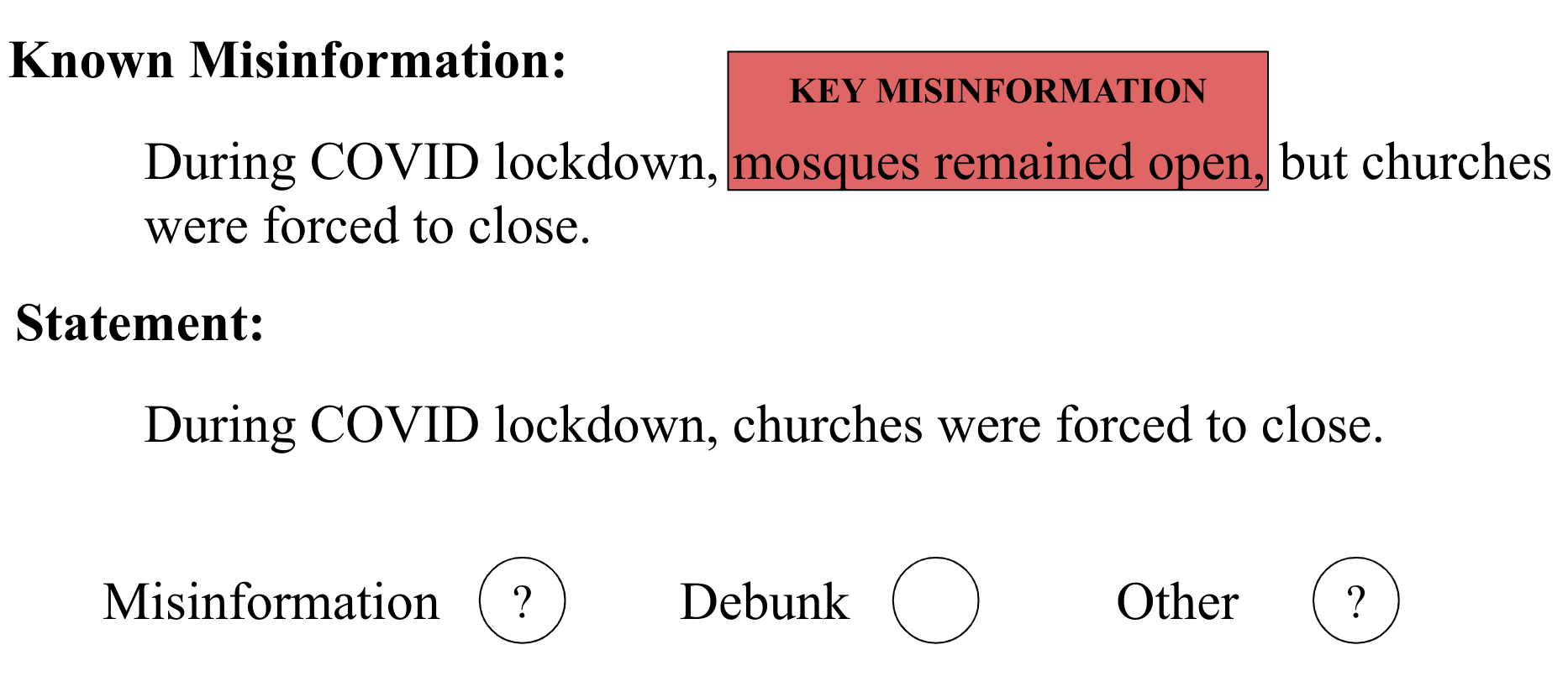}
\caption{An example of label disagreement. Without indicating that ``mosques remained open'' during COVID is the misinformation, not ``churches were forced to close'', subjectivity is introduced. The annotator may label the statement as ``misinformation'' if they thought churches closing was misinformation or ``other'' if they identify that the statement does not discuss mosques staying open.}
\label{fig:example}
\end{figure}

For the example presented in Figure~\ref{fig:example}, the known misinformation, ``During the COVID lockdown, mosques remained open, but churches were forced to close'', poses a challenge when labelling the statement ``During the COVID lockdown, churches were forced to close'' as misinformation or not. Without knowing that ``mosques remained open'' is the misinformation, it becomes difficult to make a definitive judgment. If annotation guidelines require annotators to assign a label, personal interpretations of the task may lead to disagreement.

In many cases, disagreements are resolved by a majority vote, introducing a ``gold'' label for each data point. However, such disagreements also contain valuable information and previous studies \citep{branson2010visual, Wu2023DontWA, busso2008iemocap,fayek2016modeling, zhang2019f} suggest that incorporating human disagreement or uncertainty could lead to better model performance. 

That said, these approaches often assume that annotators are reliable and that disagreements are not due to annotator mistakes. This assumption does not always hold true, especially when non-expert annotators are involved. Our experiments also demonstrate that without accounting for annotator errors, incorporating human uncertainty can actually degrade model performance.

In this work, we introduce the Efficient Annotator Reliability Assessment (EffiARA) framework, designed to evaluate the reliability of individual annotators relative to the group average. Annotator reliability serves as a proxy for estimating the error rate of each annotator and is incorporated into a weighted cross-entropy loss function. This approach allows us to adjust the importance of samples based on annotator reliability, increasing or decreasing the weight of samples accordingly, improving model training.

To apply this annotation framework, we also introduce the Russo-Ukrainian Conflict Knowledge-Based Misinformation Classification Dataset (RUC-MCD), containing debunked misinformation from EUvsDisinfo \footnote{\url{https://euvsdisinfo.eu/}} and related social media posts from X \footnote{\url{https://x.com/}} with manually annotated misinformation labels.\\

\noindent
\textbf{In summary, this study provides the following contributions:}
\begin{itemize}[noitemsep, topsep=0pt]
    \item A novel knowledge-based misinformation-detection dataset on the topic of the Russo-Ukrainian conflict.
    \item A novel annotation framework for assessing annotator reliability and maximising the number of data points per annotation.
    \item A novel approach to weighting cross-entropy loss based on annotator reliability derived from inter- and intra-annotator agreement.
    \item Baseline results on the dataset, using hard- and soft-label learning, confidence calibration \cite{Wu2023DontWA}, and annotator reliability weighted learning with TwHin-BERT-large \cite{zhang2023twhin} and Llama-3.2-1B \cite{dubey2024llama}.
    \item Open-sourced annotation framework\footnote{\url{https://github.com/MiniEggz/EffiARA}}, experimental code and dataset\footnote{\url{https://github.com/MiniEggz/ruc-misinfo}} made available on GitHub.
\end{itemize}


\section{Related Works}

Knowledge-based misinformation detection involves classifying content as misinformation based on knowledge from a trusted source.

There are two main approaches to utilising such knowledge. The first approach involves storing knowledge in a knowledge graph \cite{pan2018content} and applying graph neural networks (GNNs) to detect misinformation. The second approach keeps the knowledge in its natural language form and uses information retrieval techniques to find relevant knowledge \cite{jiang2021categorising, martin2022facter,thorne2018fever}. Once the relevant knowledge is retrieved, a classifier is applied to determine whether the claim aligns with the retrieved knowledge, contradicts it, or neither \cite{jiang2021categorising, martin2022facter,hossain2020covidlies,thorne2018fever}, thereby classifying the claim as misinformation or not.

Several datasets have been introduced to support classifier training and verification. Emergent Stance Classification \cite{ferreira2016emergent} was one of the earliest datasets, where claims were labelled based on the stance of news articles as ``for'' (supporting the claim), ``against'' (refuting the claim), or ``observing'' (neutral, without a clear stance). The Fake News Challenge \footnote{\url{http://www.fakenewschallenge.org/}} extended this idea by introducing an additional label, ``unrelated'', for articles that did not pertain to the claim.

The PHEME dataset \cite{pheme2016, derczynski2017semeval} focuses on social media rumours, classifying claims as ``Support'', ``Deny'', ``Query'', or ``Comment'' based on the provided content.

The FEVER dataset \cite{thorne2018fever} branded the task as a fact-checking task and adapted the Natural Language Inference (NLI) labels to classify text as ``Support'', ``Refuted'', or ``Not enough info'' based on the knowledge provided.

Following that, domain-specific datasets have also emerged, such as SCIFACT \cite{wadden2020fact}, which focuses on verifying scientific claims; CovidLies \cite{hossain2020covidlies} and JIANG COVID \cite{jiang2021categorising} focus on the COVID domain.

These datasets often employ multiple annotators to measure inter-annotator agreement; most datasets report relatively high levels of human disagreement. According to the kappa metric reported in previous studies, the metric varies from 0.75 (SCIFACT \cite{wadden2020fact} Cohen’s kappa) to 0.68 (\cite{thorne2018fever} Fleiss' kappa). The SCIFACT dataset, focusing on the scientific domain, achieved higher agreement, partly due to the use of domain expert annotators, whereas datasets covering broader topics often suffer from lower agreement due to the complexity and subjectivity of the task.

Disagreements are often resolved by a majority vote and apply the aggregated ``gold'' label for the classification. However, research shows the disagreements also contain highly valuable information for model training. Retaining the disagreement as a soft label rather than a single ``gold'' hard label can improve the classification performance \citep{branson2010visual, Wu2023DontWA, busso2008iemocap,fayek2016modeling, zhang2019f}, especially when the label contains some level of subjectivity \citep{khurana2024crowd, chou2024minority}. However, these approaches assume the disagreement is due to the annotator's subjectivity without considering human error. In this paper, we take human error into consideration and weight samples based on inter- and intra-annotator agreement in the soft-label training process.


\section{Problem Definition}

\noindent
\textbf{Dataset Creation} \\
Create a dataset $D$ containing a set of knowledge-based claims $C$ and social media post texts $T$ paired by relevance, forming the samples $S$ belonging to $D$. Each sample (equivalent of a data point), $S_i$, will be single-, double-, or re-annotated (annotated by one annotator once, annotated by two different annotators, or annotated by the same annotator twice), allowing the assessment of the \textit{EffiARA} (Efficient Annotator Reliability Assessment) annotation framework and the creation of a high-agreement subset of $D$ to be treated as gold-standard. Each label $L_i$ for each sample $S_i$ must be a soft label (which may be converted to a hard label), allowing for the comparison between soft- and hard-label training.\\

\noindent
\textbf{Assessing Annotator Reliability}\\
The reliability of an annotator $A^R_i$ may be determined through some combination of each annotator’s agreement with others and themselves (inter- and intra-annotator agreement). An annotation framework should be used, such that $A^R_i$ can be calculated and the number of data points $k$ is maximised.\\

\noindent
\textbf{Classifying Misinformation}\\
Our misinformation classification is to be modelled in a pairwise fashion similar to natural language inference, with post text $T_i$ being misinformation (entailing), a debunk (contradicting), or other (neutral) relative to a knowledge-based claim $C_i$ (a known piece of misinformation). A model $M$ will train to predict the label $L_i$, based on only $C_i$ and $T_i$.\\


\section{EffiARA Annotation Framework}
\label{sec:annotation-framework}

The EffiARA (Efficient Annotator Reliability Assessment) annotation framework aims to maximise the number of data points per annotation while maintaining the ability to assess annotator reliability through inter- and intra-annotator agreement. This annotation framework is well-suited for cases where there is limited access to annotators or annotations must be completed quickly. It may also be extended to much larger groups of annotators.

\begin{figure}[h]
\centering
\includegraphics[width=6cm]{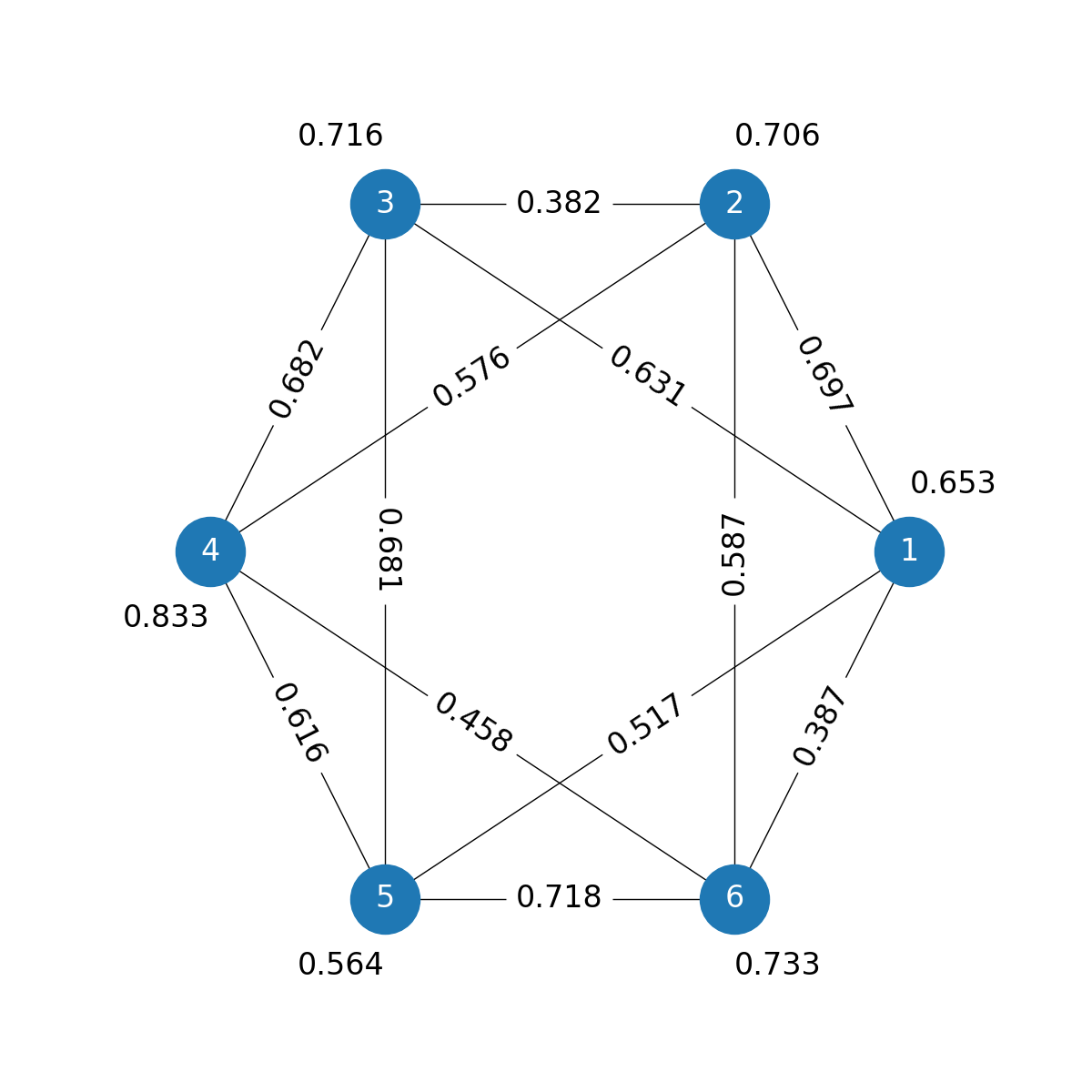}
\caption{An example of the EffiARA annotation framework with 6 annotators. Values next to each node represent an annotator's intra-annotator agreement, and values on edges represent the pairwise inter-annotator agreement. Both inter- and intra-annotator agreement were assessed using Krippendorff's alpha. Annotator reliability scores calculated weighting inter- and intra-annotator agreement equally for annotators 1-6: $0.949$, $0.989$, $1.026$, $1.108$, $0.937$, $0.992$.}
\label{fig:effiara}
\end{figure}

Figure \ref{fig:effiara} shows a graph representing the case of EffiARA with 6 annotators. Each node represents an annotator and each edge represents a set of pairwise annotations (annotations by two annotators on a set of shared samples). In the case where the number of annotators is 5 or larger, each annotator is linked to 4 other annotators, allowing for inter-annotator agreement calculations. Due to the required 4 links to other annotators, this framework should not be used where the number of annotators is less than 5. Each annotator will single-annotate and double-annotate a number of samples to increase the number of data points. Each single annotation results in a unique sample, whereas a double annotation results in one unique sample every two annotations. Single-annotated samples refer to samples annotated by one annotator only; double-annotated samples refer to samples annotated by two annotators; re-annotated samples refer to samples that have been single-annotated and then re-annotated by the same annotator. \\
\begin{algorithm}
\caption{EffiARA Sample Distribution to Annotation Projects.}
\label{alg:sample-distribution}
\begin{algorithmic}
    \STATE Let $n$ be the number of annotators
    \STATE Let $t$ be the time available for each annotator
    \STATE Let $\rho$ be the rate of annotation per unit time
    \STATE Let $k$ be the number of overall samples
    \STATE Let $d$ be the proportion of $k$ that is double-annotated in the range $0..1$
    \STATE Let $r$ be the proportion of single-annotated samples to be re-annotated in the range $0..1$
    \STATE Let $A$ be the set of annotators
    \STATE Let $D$ be the dataset containing $k$ samples

    \STATE

    \STATE \# Calculate overall samples
    \STATE $k = \left(2d + (1 + r)(1 - d)\right)^{-1} \cdot \rho \cdot t \cdot n$

    \STATE

    \FOR{$i = 1$ to $n$}
        \STATE \# Sampling performed without replacement

        \STATE \# Double-annotated samples
        \STATE $A_i \cap A_{(i+1) \bmod n} \gets \text{random\_sample}(D, \frac{dk}{2n})$
        \STATE $A_i \cap A_{(i+2) \bmod n} \gets \text{random\_sample}(D, \frac{dk}{2n})$

        \STATE \# Single-annotated samples
        \STATE $A_i \gets \text{random\_sample}(D, \frac{(1-d)k}{n})$

        \STATE \# Re-annotated samples
        \STATE $A_I \gets \text{random\_sample}(A_i, r|A_i|)$
    \ENDFOR

\end{algorithmic}
\end{algorithm}

\noindent
\textbf{Unique Samples \& Sample Distribution} \\
Algorithm \ref{alg:sample-distribution} describes how the number of samples $k$ can be calculated and distributed to the appropriate annotators. Assuming all annotators have the same amount of time, we identify 5 variables required to calculate the number of unique samples: the number of annotators, the time available for each annotator, the rate at which the average sample is annotated, the proportion of re-annotations, and the proportion of double-annotations. The ratio of time taken to produce a unique single annotation (including re-annotations) to a unique double annotation for each user is $(1+r)(d-1)$:$2d$, reflected in the equation for $k$. If a given number of samples is desired, the equation may be rearranged to find the desired number of annotators $n$, or the total time each annotator is expected to spend annotating $t$. Each parameter may be adjusted and the number of links per annotator is always $4$. Future work may investigate an alternate number of links per annotator.

\noindent
\textbf{Annotator Reliability Calculations} \\
Referring to Figure \ref{fig:effiara} may help to visualise the annotator reliability calculation process. Let $\{A_0, ..., A_n\}$ be the set of nodes representing all annotators, where $n$ is the number of annotators; let $A_x \cap A_y$ be an edge between two annotator nodes $A_x$ and $A_y$, representing the set of samples that have been double-annotated by this pair of annotators; let $a$ be a function calculating the agreement between the set of double annotations from $A_x$ and $A_y$.

A simple method for calculating each annotator’s inter-annotator agreement factor, the function $e(A_i)$, is the average of the pairwise agreement values for each edge incident to the node.

\begin{align*}
e(A_i) = & \frac{1}{|\text{Links}(A_i)|} \sum_{A_j \in \text{Links}(A_i)} a(A_i \cap A_j) \\
\text{where } & \text{the function \emph{Links} gathers all nodes}\\
& \text{linked to } A_i \text{ by an edge.}
\end{align*}

This function $e(A_i)$ can be expanded to utilise the annotator’s reliability in the inter-annotator agreement calculations. For a weighted average, the function can be written as:

\begin{align*}
e(A_i) = & \frac{1}{|\text{Links}(A_i)|} \sum_{A_j \in \text{Links}(A_i)} A_j^R a(A_i \cap A_j) \\
\text{where } & A_j^R \text{represents an annotator's reliability.}\\
\end{align*}

The aim of weighting the function is to lower the impact of ``bad annotators'' with lower than average reliability scores and raise the impact of ``good annotators'' with higher than average reliability scores in the inter-annotator agreement calculation. For this calculation, the $n$ reliability scores must be normalised around one; this can be achieved by dividing each reliability score by the average of all reliability scores, assuming no negative reliability scores. While this weighted average method may offer some improvement in calculating inter-annotator agreement, the method is suboptimal as disagreement with bad annotators is not sufficiently rewarded and disagreement with good annotators is not sufficiently punished. While out of the scope of this study, this problem may prompt improvements for future work.

Intra-annotator agreement is defined as the pairwise agreement with an annotator's own re-annotations: $a({A_i, A_{I}})$, where $A_{I}$ denotes the $i^\text{th}$ annotator's re-annotations.

With methods for calculating both inter- and intra-annotator agreement, the function to calculate annotator reliability can be introduced. This function $r(A_i, \alpha)$ can be written as:
\begin{align*}
r(A_i, \lambda) = \lambda(a(A_i \cap A_{I})) &+ (1 - \lambda)(e(A_i)) \\
\text{where } & \quad 0 \leq \lambda \leq 1.\\
\end{align*}

This reliability function can either be called once, or it can be called iteratively until reliability values converge. For the first call, each annotator's reliability is initialised to $1.0$. After each iterative step, reliability values must be normalised to have a mean value of $1.0$ for use in inter-annotator agreement function $e$.

An implementation of this framework as a Python package can be found here: \url{https://github.com/MiniEggz/EffiARA}.

\section{RUC-MCD Dataset Creation}

\subsection{Sourcing Data}
The knowledge, or known misinformation connected to the Russo-Ukrainian conflict (also known as claims), was sourced from EUvsDisinfo \footnote{https://euvsdisinfo.eu}. Each claim about the Russo-Ukrainian conflict has been fact-checked and labelled as disinformation. The social media posts were drawn from a large collection of X posts collected by our institution, hosted on an ElasticSearch \cite{elasticsearch2018elasticsearch} endpoint, enabling simple information retrieval.

\subsection{Claim-Post Pairs}
\begin{figure}[h]
\centering
\includegraphics[width=\columnwidth]{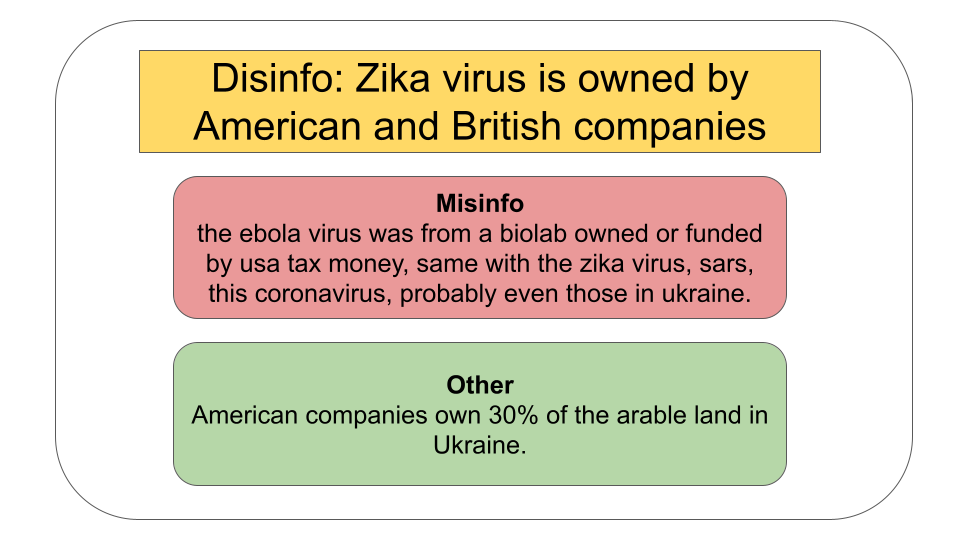}
\caption{An example of two claim post pairings using the same piece of knowledge for two social media posts. One sample is labelled ``misinfo'' and the other is labelled ``other''.}
\label{fig:claim-post-pair}
\end{figure}

With a collection of evidence and X posts, they were paired in a manner that closely follows the methodology of \citet{jiang2021categorising}. Evidence was used as the search term in an information retrieval task to find the most relevant X posts. For each claim, the 30 most relevant tweets were returned by ElasticSearch; these posts were then re-ranked using \emph{ms-marco-TinyBERT-L-2} \cite{ms-marco-TinyBERT-L-2}, with the 10 most relevant posts paired with the claim added to the dataset as 10 separate samples. An example of two claim-post pairs can be seen in Figure \ref{fig:claim-post-pair}.

Due to the large number of fact-checked claims available, a random sample of 350 was chosen from the claims originally written in English. After claim-post pairings, there were 3,500 samples available.

\subsection{Annotation}
The annotations were completed based on annotation guidelines agreed to by all annotators, 6 volunteering final-year integrated Master's students in Computer Science studying in the UK. The guidelines ask each annotator to read the evidence and post carefully and choose one of the following three labels as their primary label: ``misinfo'', ``debunk'', or ``other''. The annotator is then asked to provide a score from 1-5 indicating their confidence in the primary label, as in \citet{Wu2023DontWA}. If the confidence score is 3 or less, the annotator is asked to provide a secondary label that they believe could alternatively be assigned to the sample.

Once guidelines had been outlined and agreed upon, the samples were distributed using the EffiARA annotation framework. It was calculated that with $n = 6$, $t = 10$ (hours), $\rho = 60$ (annotations per hour), $d = 1/3$, and $r = 1/2$ that $k = (2(1/3) + 3/2 \times 2/3)^{-1} \times 60 \times 10 \times 6 = 2160$ unique samples. Once distributed, each double-annotation project contained $\frac{dk}{2n} = 80$ samples, each single-annotation project contained $\frac{(1-d)k}{n} = 240$ samples, and each re-annotation project contained $r|A_i| = 120$ samples. Initially, single- and double-annotations were completed; after completing these annotations, annotators did not annotate for two weeks before completing their re-annotations, allowing for the assessment of intra-annotator reliability. All annotators were aware of the annotation process, knowing that samples would be single-, double-, and re-annotated. Figure \ref{fig:effiara} shows the annotation graph containing agreement metrics (Krippendorff's alpha \citep{krippendorfs_alpha}) from the three-class annotation. The average Krippendorff's alpha across each double-annotation task is 0.581.

The data statistics are reported in Table \ref{tab:datasetSoftLabelDist}.

\begin{table}[ht]
    \centering
    \small
    \begin{tabular}{|c|c|c|c|c|}
        \hline
         & \multicolumn{4}{|c|}{\textbf{Label}} \\
        \hline
         \textbf{Dataset} & \textbf{Misinfo} & \textbf{Debunk} & \textbf{Other} & \textbf{Total} \\
        \hline
        High-Agreement & 80 & 29 & 447 & 556 \\
        Other & 360 & 117 & 1063 & 1540 \\
        \hline
        \textbf{Total} & 440 & 146 & 1510 & 2096 \\
        \hline
    \end{tabular}
    \caption{Distribution of hard labels in RUC-MCD, for both high-agreement and other samples, where ``other samples'' contains all single-annotations and all double-annotations where agreement criteria were not met.}
    \label{tab:datasetSoftLabelDist}
\end{table}

To form a test set for evaluating a fine-tuned model's performance on RUC-MCD, we adopt the approach from \citet{Wu2023DontWA}, using high annotator agreement as a proxy for label correctness. For a sample to be included in the high-agreement set, both annotators must have assigned the same primary label with a confidence score of 3 or higher. This high-agreement set is treated as the ``gold'' standard.

\section{Experiments}

In this study, the EffiARA reliability scores with weighted cross-entropy loss are compared with baseline classification results obtained by the same pre-trained language model and confidence calibration \cite{Wu2023DontWA}, using one encoder model and one decoder model with both soft and hard-label training; hard-label training was not possible with the confidence calibration method as it relies on the use of soft labels. Five-fold cross validation was conducted for each experimental setting; for each fold, $4/5$ of the high-agreement samples and all other samples (containing low-agreement double-annotated samples and all single-annotated samples) were used as the training set, and the final $1/5$ of the high-agreement samples used as the test set as in \citet{Wu2023DontWA}.

TwHIN-BERT-large (550M parameters) \cite{zhang2023twhin}, chosen for its pre-training on X/Twitter posts, and Llama-3.2-1B (1.23B parameters) \cite{dubey2024llama}, from the latest group of Llama models, were fine-tuned and measured for classification performance on RUC-MCD. In our fine-tuning process, we used the CLS (classification) token (for TwHIN-BERT-large) and the last token (for Llama-3.2-1B) output from the final hidden layer of the models as representations of the input. These token embeddings were then passed through an additional feed-forward neural network for classification.
These experiments were performed using the PyTorch \cite{paszke2019pytorch} and HuggingFace \cite{wolf2019huggingface} libraries. For both the BERT model and Llama model, training parameters were largely the same, following that of \citet{Wu2023DontWA}. Fine-tuning was performed for 20 epochs, a batch size of 16 for BERT and 8 for Llama (due to memory capacity), and a linear decaying learning rate of 2e-5, with 5 warm-up epochs. Without 5 warm-up epochs, the BERT models suffered from early convergence, reaching a suboptimal local minimum. This behaviour may be investigated in future research. The AdamW optimiser and cross-entropy loss, or weighted cross-entropy loss where EffiARA reliability scores were used, were chosen. For evaluation, the macro-F1 score and expected calibration error (ECE) were used as key metrics. In all experiments, the evidence was concatenated with the X post and tokenised. In total, experiments took roughly 270 GPU hours using an Nvidia Tesla A100.

\begin{table*}[h!]
\centering
\begin{tabular}{|c|c|c|c|c|}
\hline
\multirow{10}{*}{\rotatebox{90}{\textbf{BERT}}} & \multirow{5}{*}{\textbf{Hard}} & \textbf{Method} & \textbf{F1-Macro} & \textbf{ECE} \\ \cline{3-5}
& & Baseline            & \textbf{0.699} (0.05) & 0.151 (0.02) \\ \cline{3-5}
& & EffiARA (inter)      & \underline{0.698} (0.05) & \underline{0.144} (0.02) \\ \cline{3-5}
& & EffiARA (intra)      & 0.690 (0.07) & 0.152 (0.03) \\ \cline{3-5}
& & EffiARA (inter+intra)& 0.677 (0.12) & \textbf{0.119} (0.03) \\ \cline{2-5}

& \multirow{5}{*}{\textbf{Soft}} & Baseline            & 0.691 (0.07) & \textbf{0.071} (0.01) \\ \cline{3-5}
& & EffiARA (inter)      & \underline{0.728} (0.04) & \underline{0.072} (0.02) \\ \cline{3-5}
& & EffiARA (intra)      & 0.722 (0.07) & 0.079 (0.01) \\ \cline{3-5}
& & EffiARA (inter+intra)& \textbf{0.740} (0.06) & 0.077 (0.02) \\ \cline{3-5}
& & Confidence Calibration & 0.627 (0.03) & 0.116 (0.01) \\ \hline

\multirow{10}{*}{\rotatebox{90}{\textbf{Llama}}} & \multirow{5}{*}{\textbf{Hard}} & \textbf{Method} & \textbf{F1-Macro} & \textbf{ECE} \\ \cline{3-5}
& & Baseline            & \underline{0.738} (0.02) & 0.116 (0.01) \\ \cline{3-5}
& & EffiARA (inter)      & 0.726 (0.04) & 0.121 (0.02) \\ \cline{3-5}
& & EffiARA (intra)      & \textbf{0.751} (0.05) & \textbf{0.106} (0.02) \\ \cline{3-5}
& & EffiARA (inter+intra)& 0.726 (0.07) & \underline{0.111} (0.02) \\ \cline{2-5}

& \multirow{5}{*}{\textbf{Soft}} & Baseline            & 0.730 (0.09) & 0.093 (0.02) \\ \cline{3-5}
& & EffiARA (inter)      & 0.724 (0.06) & 0.094 (0.01) \\ \cline{3-5}
& & EffiARA (intra)      & \underline{0.732} (0.07) & \textbf{0.079} (0.01) \\ \cline{3-5}
& & EffiARA (inter+intra)& \textbf{0.756} (0.07) & \underline{0.092} (0.01) \\ \cline{3-5}
& & Confidence Calibration & 0.638 (0.07) & 0.124 (0.01) \\ \hline

\end{tabular}
\caption{Classification performance and model calibration of BERT and Llama models with hard- and soft-label training, using EffiARA reliability scores and confidence calibration. For each model and label type training, a baseline is provided, using unweighted cross-entropy loss and no calibration of confidence. For each model and label type, bold indicates the highest-performing method and underlined indicates the second-highest-performing method.}
\label{table:final_res}
\end{table*}

Due to label imbalance, we combine the labels ``other'' and ``debunk'', leaving only ``misinfo'' and ``other'' in our experiment. This study focuses on the detection of misinformation; the fine-grained classification of other classes is not essential. All data was collected using three-label classification, with the ``debunk'' and ``other'' labels only merged as a data-preprocessing step in our experiments. Inter- and intra-annotator agreement is calculated after this step, resulting in marginally different Krippendorff's alpha metrics, going from 0.578 with three labels to 0.592 with two labels for inter-annotator agreement and from 0.701 to 0.707 for intra-annotator agreement.

\subsection{Label weight and Soft Label Generation}
The annotator reliability metric aims to capture how ``good'' an annotator is, using their inter- and intra-annotator agreement scores, calculated as described in Section \ref{sec:annotation-framework}. These reliability measures were then used in the weighted cross-entropy loss function, increasing the importance of correctly classifying ``better'' annotators in the model training process. As well as weighting the loss function, reliability scores are also used to aggregate double annotations to create soft labels, using a weighted average. This study uses Krippendorff's alpha \cite{krippendorff1970estimating, hayes2007answering, castro-2017-fast-krippendorff} as the agreement metric in all inter- and intra-annotator agreement calculations.

Three settings were used to calculate annotator reliability: inter-annotator agreement only, intra-annotator agreement only, and a 50:50 weighting of inter- and intra-annotator agreement. The reliability scores can be seen in Table \ref{table:annotator_reliability}.

\begin{table}[h!]
\centering
\resizebox{\columnwidth}{!}{%
\begin{tabular}{l|cccccc}
\toprule
& \textbf{$A_1$} & \textbf{$A_2$} & \textbf{$A_3$} & \textbf{$A_4$} & \textbf{$A_5$} & \textbf{$A_6$} \\
\midrule
\textbf{Inter}       & 1.071 & 0.962 & 1.011 & 0.891 & 1.119 & 0.945 \\
\textbf{Intra}       & 0.960 & 1.000 & 0.990 & 1.157 & 0.824 & 1.068 \\
\textbf{Both}        & 1.011 & 0.986 & 1.000 & 1.038 & 0.959 & 1.007 \\
\bottomrule
\end{tabular}%
}
\caption{Annotator reliability scores for each annotator in each reliability calculation setting once labels had been reduced to ``misinfo'' and ``other''.}
\label{table:annotator_reliability}
\end{table}

To generate the soft label, the confidence in the primary label is converted to a value between $0$ and $1$. This conversion uses Equation \ref{eq:confidence-conversion}, where $n$ is the number of classes, $C$ is the confidence in the primary label, and $\textit{MaxC}$ is the maximum confidence value. In this study, these values are $n = 2$, $\textit{MaxC} = 5$, and $C \in \{1, 2, 3, 4, 5\}$.

\begin{equation}
\label{eq:confidence-conversion}
P = \frac{1}{n} + \frac{n-1}{n} \cdot \frac{C-1}{{\text{MaxC}} - 1}
\end{equation}

This confidence score is assigned as the probability that the sample belongs to the class of the primary label. If the number of classes is greater than two, if a secondary label has been selected, it is assigned the probability $\min(P, 1-P)$ to ensure the secondary label probability is not higher than that of the primary label. The remaining probability is distributed evenly amongst the remaining classes. With binary classification, the primary label is assigned probability $P$ and the secondary label $1-P$. To generate the soft label for a double-annotated sample, a mean of the two annotators' individual soft labels is taken, weighted using the annotators' reliability scores. 

This study uses both hard- and soft-label learning. By first creating the soft label, it can be transformed to a hard label using an argmax function for use in hard-label training experiments and for the test set.

\subsection{Confidence Calibration}
To assess the performance of the EffiARA annotator reliability calculations, classification results were compared to those obtained using the Bayesian confidence calibration method introduced by \citet{Wu2023DontWA}. After confidence calibration, soft labels are generated as previously described.

\section{Results \& Discussion}
Table \ref{table:final_res} presents the classification performance and model calibration results achieved in this study.

\textbf{Cross-entropy loss weighted by EffiARA reliability scores offers significant advantages.} Treating vanilla hard-label training as the baseline, annotator-reliability-based sample weighting offers improvements of up to 0.041 and 0.018 in macro-F1 score for BERT and Llama respectively, showing the effectiveness of the method on both encoder and decoder models. For both Llama and BERT, the highest performing setting was annotator-reliability-based sample weighting utilising both inter- and intra-annotator agreement to establish annotator reliability. This method was more effective when applied to soft-label learning for both BERT and Llama.

For both BERT and Llama, statistical significance was assessed, comparing the highest performing baseline and the highest performing result using cross-entropy loss weighted by annotator reliability scores using paired t-tests \citep{student1908probable} due to the approximately normally distributed differences in results; the Shapiro-Wilk test \citep{shapiro1965analysis} was used to test whether the data was normally distributed. For BERT, comparing the hard-label baseline with the highest performing method resulted in a P-value of $0.008 < 0.05$, meaning the results were statistically significant. For Llama, again comparing the hard-label baseline and the highest performing method, a P-value of $0.675 > 0.05$ was obtained, meaning the results were not statistically significant, despite offering some improvement. Further investigation, particularly through the introduction of more datasets, is required to understand this.

\textbf{EffiARA performs well with low-to-moderate-agreement data.} The average inter-annotator agreement in double-annotated samples within RUC-MCD with two classes is 0.592, which is low-to-moderate agreement \cite{krippendorff2019content}. With crowdsourcing often containing disagreement \cite{dumitrache2018crowdtruth}, it could make EffiARA a viable approach to mitigate against poor annotators. Future work may investigate the performance of EffiARA in expert annotation scenarios and crowdsourcing, where the number of annotators is much higher.

\textbf{Confidence calibration is ineffective when applied to pairwise annotations.} Bayesian confidence calibration was consistently the worst performing experimental setting in both classification performance and model calibration. This was expected as confidence calibration relies on higher annotator overlap. EffiARA's pairwise annotation method is not compatible with confidence calibration, justifying the requirement for another approach, such as annotator-reliability-based sample weighting.

\textbf{Soft- and hard-label training results vary.} While the highest performing models are trained on soft labels using the annotator-reliability-based sample weighting, hard-label training performs better than soft-label training for both BERT and Llama baselines. Expected calibration error is significantly improved through the use of soft labels though, indicating that the model's confidence in its output is more accurate with soft-label training.

\section{Conclusion}
This study has introduced the EffiARA annotation framework, shown to increase knowledge-based classification performance over baseline hard- and soft-label training for a BERT and Llama model through annotator-reliability-based sample weighting. The combination of inter- and intra-annotator agreement was shown to be the highest performing with a macro-F1 score of 0.756 for Llama-3.2-1B and 0.740 for TwHIN-BERT-large, outperforming baselines by 0.018 and 0.041 respectively. While only 6 annotators were used in this study, this framework can be expanded to use any number of annotators, meaning it is applicable to crowdsourcing; further studies may investigate this application of the EffiARA annotation framework and annotator-reliability-based sample weighting. The newly created dataset, RUC-MCD, has also been made publicly available.

\section{Limitations}

This preliminary study introducing EffiARA shows that the annotation framework is a viable approach to obtaining a large number of data points when annotator time is scarce while maintaining the ability to assess inter- and intra-annotator agreement. There is, however, scope for further investigation into the framework's optimisation and effectiveness in other settings. Avenues of future research may include testing the effect of different hyperparameters controlling the proportion of re-annotated and double-annotated data points, the proficiency of the framework with different numbers of annotators, and its performance on different datasets and supervised learning tasks. Without testing on a number of datasets, it is unclear how well the EffiARA annotation framework will generalise and perform in other circumstances.

This work relies on fact-checking organisations to always be a reliable source of truth. Placing so much trust in one organisation may be dangerous in some cases but verifying the work of fact-checking organisations is beyond the scope of this study. For use in industry, however, it is important that all the evidence itself is trustworthy. For any organisation that may want to employ misinformation detection technologies on their social media platform, it is likely suitable that the evidence base is maintained manually.

The detection of misinformation can often be reliant on temporal information. For example, if somebody claimed that ``Person X did Y'' before that person did action Y, it would be misinformation. The same statement made a month later, after Person X did Y, would no longer be misinformation. Including this temporal dimension in the misinformation classification task was beyond the scope of this study but it is an important edge case to consider, should this technology be used to automatically detect misinformation on social media. 

The application of misinformation detection technology must be carefully considered as the misidentification of it in public settings may sway opinions and narratives, causing unintended consequences.

\section{Ethics}
This study employed six annotators who all signed consent forms, agreeing to their annotations being used for this research and acknowledging and accepting the risk of reading offensive or upsetting social media posts. Any X post data, potentially containing personal information, has been omitted from the publicly available dataset; only the post ID is stored, enabling users to only use posts that remain public. This study was approved by the Ethics Board of the School of Computer Science, The University of Sheffield.

\bibliography{custom}

\end{document}